\documentclass{article}
\usepackage[final,nonatbib]{neurips_2023}

\usepackage[utf8]{inputenc} %
\usepackage[T1]{fontenc}    %
\usepackage[pdfborder={0 0 0}]{hyperref}       %
\usepackage{url}            %
\usepackage{booktabs}       %
\usepackage{amsfonts}       %
\usepackage{nicefrac}       %
\usepackage{microtype}      %
\usepackage{xcolor}         %

\newif\ifpreface
\prefacetrue %

\newif\ifabstract
\abstracttrue %

\newif\ifcontent
\contenttrue %
\makeatletter
\renewcommand{\@noticestring}{Preprint version, see \url{https://doi.org/10.1515/9783110785944-005}}
\makeatother

\input{preamble.tex}

\title{Sparse Partitioning Around Medoids}

\author{%
  Lars Lenssen\\
  TU Dortmund University\\
  \texttt{lars.lenssen@tu-dortmund.de} \\
  \And
  Erich Schubert\\
  TU Dortmund University\\
  \texttt{erich.schubert@tu-dortmund.de} \\
}

\begin{document}
\maketitle

\bgroup
\let\subsubsection\subsection
\let\subsection\section
\let\tsum\undefined\newcommand{\tsum}{\textstyle\sum}
\let\TD\undefined\newcommand{\TD}{\ensuremath{\mathrm{TD}}}
\let\TL\undefined\newcommand{\TL}{\ensuremath{\mathrm{\ell}}}
\let\Var\undefined\newcommand{\Var}{\operatorname{Var}}
\let\norm\undefined\newcommand{\norm}[1]{\left\lVert#1\right\rVert}
\let\snorm\undefined\newcommand{\snorm}[1]{\left\lVert\smash{#1}\mathstrut\right\rVert}
\let\nearest\undefined\newcommand{\nearest}{{n_1}}%
\let\second\undefined\newcommand{\second}{{n_2}}%
\xdefinecolor{Pa1}{RGB}{132, 184, 24} %
\xdefinecolor{Pa2}{RGB}{241, 135, 0} %
\xdefinecolor{Pa3}{RGB}{0, 155, 170} %
\xdefinecolor{Pa4}{RGB}{191, 2, 127} %
\xdefinecolor{Pa5}{RGB}{250, 220, 0} %
\xdefinecolor{Pa6}{RGB}{228, 103, 8} %
\newcommand{\showmark}[2][]{\bgroup\begin{tikzpicture}[baseline=-.6ex]
\node[inner sep=.5ex,mark size=0.5ex,#1]{\nullfont\pgfuseplotmark{#2}};
\end{tikzpicture}\egroup}
\pgfplotsset{width=7cm,compat=1.3}

\ifabstract
\begin{abstract}
\index{Partitioning Around Medoids (PAM)}
\index{k-medoids clustering}
\index{Facility Location Problem}
Partitioning Around Medoids (PAM, $k$-Medoids) is a popular clustering
technique to use with arbitrary distance functions or similarities,
where each cluster is represented by its most central object,
called the medoid or the discrete median.\index{Median}
In operations research, this family of problems is also known
as facility location problem (FLP).
FastPAM recently introduced a speedup for large~$k$  to make it applicable for larger
problems, but the method still has a runtime quadratic in~$N$.
In this chapter, we discuss a \emph{sparse and asymmetric} variant of
this problem, to be used for example on graph data such as road networks.
By exploiting sparsity, we can avoid the quadratic runtime and memory
requirements, and make this method scalable to even larger problems,
as long as we are able to build a small enough graph of sufficient connectivity to perform local optimization.
Furthermore, we consider asymmetric cases, where the set of medoids is not identical
to the set of points to be covered (or in the interpretation of facility location,
where the possible facility locations are not identical to the consumer locations).
Because of sparsity, it may be impossible to cover all points with just $k$ medoids
for too small $k$, which would render the problem unsolvable, and this breaks 
common heuristics for finding a good starting condition.
We, hence, consider determining $k$ as a part of the optimization problem
and propose to first construct a greedy initial solution with a larger $k$,
then to optimize the problem by alternating between PAM-style ``swap'' operations
where the result is improved by replacing medoids with better alternatives
and ``remove'' operations to reduce the number of $k$ until neither allows
further improving the result quality.
We demonstrate the usefulness of this method on a problem from
electrical engineering, with the input graph derived from cartographic data.

\end{abstract}
\fi

\textcolor{red}{This is a preprint of Lars Lenssen and Erich Schubert. ``5.1 Sparse Partitioning Around Medoids''.
In: Machine Learning under Resource Constraints -- Fundamentals (Volume 1) edited by Katharina Morik and Peter Marwedel, Berlin, Boston: De Gruyter, 2023, pp.{} 182-196.
\url{https://doi.org/10.1515/9783110785944-005}}

\ifcontent
\subsection{Introduction}

The algorithm Partition Around Medoids (PAM, \cite{Kaufman/Rousseeuw/87a,Kaufman/Rousseeuw/90b}),
also known as $k$-medoids, is a
popular clustering algorithm used as alternative to $k$-means clustering\index{k-means}
when one wants to minimize other distances than squared errors distance.
Similar to $k$-means, it aims at minimizing the sum of distances from a
cluster center, but the cluster center in $k$-medoids is one of the data points
and called a medoid, and the distance function here may be arbitrary.
This increases the flexibility over $k$-means, which uses the arithmetic mean
as cluster center: the mean minimizes squared errors, and because of this
$k$-means only minimizes Bregman divergences such as the squared Euclidean distance.
Even on one-dimensional data it does not minimize the linear error, as easily
seen from the difference between the arithmetic mean and the median.
While $k$-means minimizes the sum-of-squared errors, $k$-medoids, with
$k$~representative medoids $m_i$ minimizes the absolute error criterion
(``total deviation'', TD):
\begin{equation}
    \TD := \sum\nolimits_{i=1}^{k} \sum\nolimits_{x_c \in C_i} d(x_c,m_i) 
\end{equation}
where $d(x_c,m_i)$ is the distance between data point~$x_c$ of cluster~$C_i$ and medoid~$m_i$; not necessarily the Euclidean distance,\index{Euclidean distance}
and not necessarily a metric.
The difference between the arithmetic mean, the per-axis median,
the geometric median, and the medoid of a data set is exemplified
in Figure~\ref{fig:medoid}. It can be seen that the medoid is less
sensitive to outliers than the arithmetic mean, and also that $k$-means
does not minimize Euclidean distances (but the squared distances).

\begin{figure}\centering
\begin{minipage}{.4\textwidth}\centering
\begin{tikzpicture}[scale=2]
\tikzstyle{p}=[minimum size=.8ex, inner sep=0, fill=black, circle]

\draw[black!50,densely dotted, thick] (-1.05,-1.05) rectangle (1.05, 1.05);

\node[p] (p0) at (1.000000, -0.426151) {};
\node[p] (p1) at (0.375152, 1.000000) {};
\node[p] (p2) at (0.050578, -0.045320) {};
\node[p] (p3) at (-0.734840, -0.010723) {};
\node[p] (p4) at (-1.000000, -0.422913) {};
\node[p] (p5) at (-0.892265, -0.632368) {};
\node[p] (p6) at (-0.314216, -1.000000) {};
\node[p] (p7) at (0.174161, -0.886315) {};
\node[p] (p8) at (-0.500000, -0.500000) {};

\foreach \x in {0,...,8} { \draw[very thick,Pa1,opacity=.5,->,>=latex] (p\x) -- (p8); }

\node[Pa2] (mean) at (-0.204603, -0.324866) {\pgfuseplotmark{square*}};
\node[Pa3,mark size=.6ex] (median) at (-0.314216, -0.426151) {\pgfuseplotmark{triangle*}};
\node[Pa4,mark size=.7ex] (gmedian) at (-0.442589, -0.461613) {\pgfuseplotmark{diamond*}};
\node[Pa1,mark size=.6ex, very thick] (medoid) at (-0.500000, -0.500000) {\pgfuseplotmark{o}};

\end{tikzpicture}
\end{minipage}
\hfill
\begin{minipage}{.58\textwidth}\centering\footnotesize

Euclidean ($L_2$), Manhattan ($L_1$) and\\
squared Euclidean ($L_2^2$) distance sums\\
to the different central points.

\setlength{\tabcolsep}{2pt}
\begin{tabular}{lccc}
&$L_2$
&$L_2^2$
&$L_1$
\\
\textcolor{Pa2}{\showmark{square*}\enskip arithmetic mean}
& 6.909 &{\bf 6.330 }& 8.674 \\
\textcolor{Pa3}{\showmark[mark size=0.6ex]{triangle*}\enskip per-axis median}
& 6.761 & 6.530 &{\bf 8.267 }\\
\textcolor{Pa4}{\showmark[mark size=0.6ex]{diamond*}\enskip geometric median}
&{\bf 6.712 }& 7.008 & 8.431 \\
\textcolor{Pa1}{\showmark[very thick]{o}\enskip Euclidean medoid}
& 6.726 & 7.391 & 8.526 \\
\end{tabular}
\end{minipage}

\caption[Four different central points: the arithmetic mean,
per-axis median, geometric median, and the Euclidean medoid]{
Four different central points:
the arithmetic mean \textcolor{Pa2}{\showmark{square*}},
per-axis median \textcolor{Pa3}{\showmark[mark size=0.6ex]{triangle*}},
geometric median \textcolor{Pa4}{\showmark[mark size=0.6ex]{diamond*}},
and the Euclidean medoid \textcolor{Pa1}{\showmark[very thick]{o}}.}
\label{fig:medoid}
\end{figure}
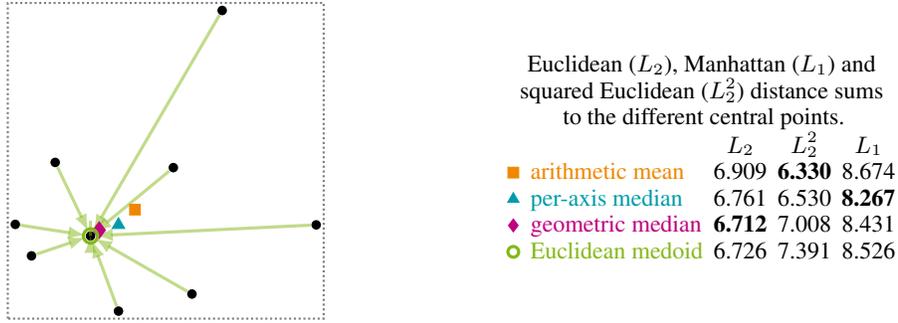

In operations research, the $k$-medoids problem is also known as the
(discrete) facility location problem. Several variants of this problem have been researched there.
The variants differ mainly in the objective function to be minimized.
For example, $k$-center instead minimizes the maximum distance of all points to their assigned cluster centers.
There has been substantial research in the area of finding approximation algorithms for all these different problems. 

Unfortunately, the algorithms commonly used for $k$-medoids are not very
scalable to large problems, as we will discuss next.

\subsection{Runtime Complexity of Partition Around Medoids}
\label{sec:lenssen-pam-complexity}

The $k$-medoids problem is NP-hard~\cite{Kariv/Hakimi/79a}, hence we have to resort
to approximate solutions, using greedy and local optimization techniques.
The PAM algorithm is such an approach: its initialization (known as BUILD)
is a greedy approximation to the $k$-medoids problem, which afterwards is
refined using a local search (called SWAP). Greedy initialization chooses $k$
times the point which reduces the error the most; local search then optimizes
this solution by searching for the best way to swap one of the cluster centers
with a non-center. While the name $k$-medoids resembles $k$-means, the standard
PAM algorithm works different from the standard $k$-means algorithm.
A $k$-means like strategy of alternating optimization for $k$-medoids has been
proposed several times \cite{Maranzana/63a,Hastie/etal/2001a,Reynolds/etal/2006a,Park/Jun/2009a},
but was shown to produce worse solutions than a swap-based approach such as
PAM \cite{Teitz/Bart/68a,Rosing/etal/79a,Schubert/Rousseeuw/2021a}.
\textcite{Kanungo/etal/2004a} proposed a swap-based approach to also improve the
results of $k$-means, but it is rather expensive as we will see next.

Both the greedy initialization as well as the local search require all
pairwise distances stored in a distance matrix. Greedy initialization performs
$k$ iterations, each of cost $O(N^2)$ to find the best medoid to add.
PAM's swap evaluates $O(k(N-k))$ potential swaps, each with a reduced effort of $O(N-k)$
operations by computing only the change in the loss function.
Hence each swap takes $O(k(N-k)^2)$ time to find, which already
was an improvement over the naive approach in $O(k^2(N-k)^2)$.
The resulting runtime complexity of PAM is $O(kN^2i)$, where $i$ is
the number of iterations until convergence for which little is known except that
it usually is reasonably small, and likely has an unfavorably high worst case
just as with $k$-means.

We have recently proposed improved versions of PAM named
FastPAM~\cite{Schubert/Rousseeuw/2019a} and
FasterPAM~\cite{Schubert/Rousseeuw/2021a}, which provide a substantial
speedup over PAM by eliminating the nested loop over the $k$ medoids.
By furthermore greedily performing the first swap that improves the loss
(instead of the best swap) and random initialization, this allowed us to
decrease the runtime complexity to $O(N^2i)$ with an empirically much
smaller $i$ (but supposedly a similar theoretical worst case).

Because both methods use each pairwise distance several times---and the
method is in particular interesting to use with more complex and hence
expensive distance function---it is prohibitive to not use it with a
pairwise distance matrix. Hence both methods also require $O(N^2)$ memory.

\subsection{Sparse Partitioning Around Medoids}
\label{sec:lenssen-sparsepam}

A large part of these pairwise distances
may be unnecessary to know exactly. It is easy to see that given some
assignment of points to medoids, and the maximum distance $\tau$ of this
\emph{assignment}, we could replace all values larger than $\tau$ in this
\emph{input} distance matrix with $\tau$, and the solution would not change.
Hence there is some natural \enquote{cut-off} to distances,
and larger values do not contribute to the solution.
If our distance function satisfies the triangle inequality,
we may be able to omit computing some of these large distances
(e.g., with the algorithm of \textcite{Newling/Fleuret/2017a}).

In this research, we want to focus on a different scenario,
where the cut-off may be given in advance (and may be different for
each point), but the distance is not necessarily metric.
An real-word example for such as problem will be introduced
in Section~\ref{sec:lenssen-usecase}.
While we can (and, effectively, will) treat distances considered uninteresting
for the application
as infinite (or sufficiently large) values, using a sparse\index{sparse data} storage
or the distances only immediately reduces the memory usage, not the runtime.
Unfortunately, this also easily breaks the optimization procedure,
which relies on first finding a \emph{feasible} initial solution,
then performing local changes that \emph{improve} the solution.
A greedy strategy such as the one discussed above is usually not able
to find a valid initial solution for a small $k$ (and in particular, for a
very small $k$ the problem may become unsatisfiable with a finite loss).
In such cases, the local optimization will also not help, as neighbor
solutions will often still be invalid, and hence make no progress.
This is most easily seen if the data set consists of many components
that are not connected with edges of finite length.

Instead of searching directly for a solution with $k$ centers,
we can solve a second problem
of $k$-medoids clustering at the same time: how should we choose $k$?
Already with $k$-means clustering, choosing the \enquote{optimal}~$k$
has shown elusive to a general solution, and is mostly performed by some
crude heuristic such as the infamous Elbow criterion, which is
frequently misused.

If we allow the algorithm to vary $k$, we can much more easily find a
valid initial solution (e.g., by choosing the best unconnected vertex
until everything is covered). But of course this will usually yield
a much higher number of clusters $k$ than desired. But if we perform
a multi-criteria optimization in the refinement phase, we may be able
to reduce the number of clusters along with minimizing our main objective.

When varying $k$, we will obtain a Pareto front of solutions that are
all optimal in one way or another. This can be formalized as solutions
not \enquote{dominated} by any other solution in each criterion at the
same time.
To reduce the set of remaining candidate solutions, it is best if we
have some additional constraints to satisfy based on the particular
problem to solve.

\subsection{Use Case: Simulation of Electrical Substation}
\label{sec:lenssen-usecase}

We obtain networks using
OSMOGrid, which implements ideas of distribution network generation of Kays et al.~\cite{Kays/etal/2017a}
on the basis of public data (OpenStreetMap, OSM).\index{OpenStreetMap}
The electrical grid is modeled to follow the streets, and the buildings are used to model consumers.
Power consumption is estimated based on zoning and building size, and used to simulate the load flow in the grid.
We have made some graph simplifications in preparation for the problems presented below.
We remove dead ends, and move the consumers locations (i.e., buildings)
to the next point in the street network.
Figure~\ref{sec:lenssen-usecase-fig1} shows the simulation based on the township Witten Stockum.

\begin{figure}[b!]
	\center
	\includegraphics[width=1.0\textwidth]{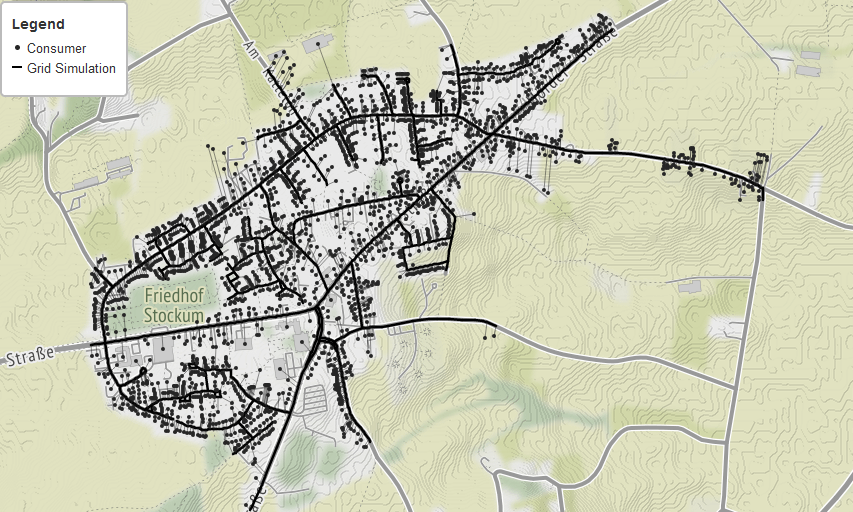}
	\caption{Simulation of an electrical grid based on OSM data of Witten Stockum.}
	\label{sec:lenssen-usecase-fig1}
\end{figure} 

On the basis of this graph structure,\index{graph data} there are different computational tasks in which resource-efficient clustering models are necessary.
One of these tasks is the simulation of electrical substations within the graph.
We want to identify the optimal positions of power substations, such the electric losses in the network are minimized.
As the electric loss is related to load, voltage, and the cable length,
we approximate it using the distance between substations and their connected consumers, which is weighted by the load of the consumers.
We describe this as a facility location problem, which comes from urban and public service planning.
The objective function $\text{FL}$ for facilities and demand points is
\begin{equation}
    \text{FL} =  \sum\nolimits_{i\in \text{Demands}} d(i,m(i)) + \sum\nolimits_{j\in \text{Centers}} c(j)
    \quad,
\end{equation}
with $c(j)$ as the cost of opening a facility,
and $d(i,m(i))$ as the distance between consumer~$i$ and the assigned center~$m(i)$.
$\text{FL}$ has strong similarities to the objective function of $k$-medoids.
We take the facilities as the electrical substations and the consumers as the demand points.
Figure~\ref{sec:lenssen-usecase-fig2} shows results of clustering the consumers with FasterPAM
for $k=4$ substations for the generated graph for Witten Stockum.
We can observe that cluster assignment follows the road network,\index{Road network} and consumers
are not necessarily assigned to the closest center ``as the bird flies''.

\begin{figure}[t!]
	\center
	\includegraphics[width=1.0\textwidth]{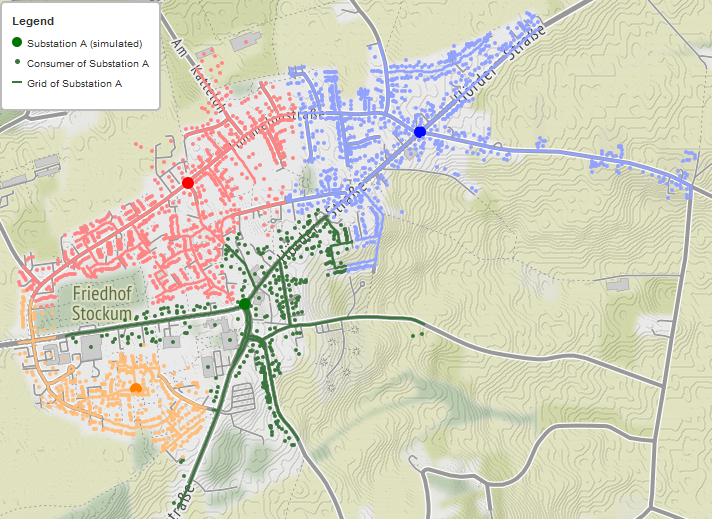}
	\caption{Clustering of the demand points of the generated graph structure according to optimal substation locations with $k=4$ using FasterPAM.}
	\label{sec:lenssen-usecase-fig2}
\end{figure} 

Even with the FasterPAM improvements, the runtime complexity is $O(N^2i)$
for $N$~nodes in the graph and $i$~iterations of the optimization procedure.
The underlying OSM planet file contains about 1.2 TB of data.
Even though we are only interested in modeling smaller areas of the world,
we need to achieve reduction in complexity for solving the task
for whole cities or regions in an acceptable runtime, as these will nevertheless
contain several thousands of houses.
We take advantage of some properties of a typical electrical network.
We only consider nodes with
at least 3 outgoing edges as possible optimal substation locations (except for disconnected points). The optimal position on a single edge is trivial to calculate and is neglected.
Hence, it is beneficial to formulate this as an asymmetric problem, where demand points and facility locations are no longer the same set.
The distance matrix then no longer has to be calculated for all node pairs,
but only for all demand points and substation location connections.
This reduces the complexity to $O(Nmi)$ for $N$ consumers, $m$ possible substation locations, with $m<N$.
If we further limit the maximum distance between a consumer and a substation (to limit the power losses),
this distance matrix becomes \emph{sparse}, i.e., we now have missing values that we can consider as infinite values.
If we do not store these missing values and iterate using appropriate sparse data structures,
we expect to further reduce the runtime to $O((e+N+m)\cdot i)$ for $e$ edges.
Assuming a similar density of houses and roads everywhere, we can expect the number of edges $e$ to be
approximately linear in the \emph{area} of the map we are processing.  

\subsection{Sparse \boldmath$k$-Medoids}
To use $k$-medoids clustering for problems with asymmetric and sparse input data,
we have to adapt the objective function of $k$-medoids.
We still want to minimize the ``total deviation'' of all data points $\{x_1,...,x_N\}$
from the current set of medoids $M\subseteq \{y_1,...,y_m\}$,
but we no longer assume $M\subset X$ as in tranditional $k$-medoids.
Furthermore, for some points, there currently may be no closest reachable medoid $m(x_i)$,
and all distances from this $x_i$ to all medoids $m_i\in M$ are undefined.
In such cases, we have to incorporate a penalty $\pi(x_i)$ in our loss $\TL$:
\index{Loss}
\begin{equation}
    \TL := \sum\nolimits_{i=1}^N \begin{cases}
    \pi(i) & \text{if } m(x_i)=\text{undefined}\\
    d(x_i,m(x_i)) & \text{otherwise}
    \end{cases}
\enskip . %
\end{equation}
Note that we allow the set $M$ to change in size below.
The penalty $\pi(i)$ can be used to trade the loss of not covering all possible
data points against having larger distances. We do not further consider tuning this parameter
below, but we instead use $\pi(i) = \pi = \text{const} \rightarrow \infty$ to enforce a complete
coverage. Because such extreme values can cause numerical problems, our implementation
always uses pairs $(i,d)$ to store a loss (and a loss change):
an integer $i$ to count the number of unassigned points, and the sum of distances of assigned points $d$,
such that mathematically we have $\TL = i\cdot \pi + d$, but do not suffer from numerical problems.

Based on the objective function, we introduce DynBUILD (Dynamic Asymmetric BUILD initialization)
as an adaptation of the BUILD algorithm of \textcite{Kaufman/Rousseeuw/87a,Kaufman/Rousseeuw/90b}
to asymmetric sparse input datasets. %
The greedy BUILD approach is supplemented by a dynamic increase of $k$,
if after choosing $k$ medoids, some objects (still) are not reachable by the current set of medoids.
The algorithm hence always choses at least $k$ medoids and covers all consumers. %
As a baseline, the strategy denoted Random simply uses a given percentage of points as initial cluster centers,
and may hence yield an initial solution where constraints are violated,
but our improved DynSWAP procedure will repair these while optimizing the assignment.
Sparse++ is an adaptation of the well-known $k$-means++~\cite{Arthur/Vassilvitskii/2007a} method to sparse data,
where cluster centers are chosen proportionally to how many points they cover (again, we continue choosing additional
centers until all constraints are satisfied).

\begin{algorithm2e}[b!]
\SetAlgoLined\DontPrintSemicolon\SetArgSty{textrm} %
\SetKw{KwAnd}{and}\SetKw{KwOr}{or}
\SetKw{KwIf}{if}\SetKw{KwElse}{else}
\caption{DynBUILD: Dynamic Asymmetric BUILD initialization}
$\TL,M \gets (\infty,\infty),\emptyset$\;
\tcc{Choose the first medoid:}
\ForEach(\tcp*[f]{compute loss for each $y_j$}){$y_j$}{ 
    $\TL_j\gets (\sum_i \pi(i),0)$\tcp*[r]{everything is unassigned}
    \ForEach(\tcp*[f]{check neighbors (sparse)}){$x_o\in N(y_j)$} {
		$\TL_j \gets \TL_j + (-\pi(o),d(x_0,y_j))$\;
	}
	\lIf(\tcp*[f]{current best}){$\TL_j < \TL)$} {
		$\TL,M\gets \TL_j,\{y_j\}$ 
    }
}
\tcc{Choose the remaining medoids:}
\For{$i=1\ldots k-1$} {
	$\Delta\TL^*,y^* \gets (0,0),\emptyset$\tcp*[r]{storage for best solution}
	\ForEach {$y_j\notin M$} {
		$\Delta\TL_j \gets (0,0)$\tcp*[r]{loss change accumulator}
		\ForEach(\tcp*[f]{check neighbors (sparse)}){$x_o\in N(y_j)$} {
			$\delta\pi \gets -\pi(o) $ \KwIf $d_\nearest(o)=\infty$ \KwElse $0$ \;
			$\delta d \gets d(x_o,y_j)-d_\nearest(o)$\;
			\lIf {$\delta\pi<0 $ \KwOr $\delta d<0$} {
				$\Delta\TL_{j}\gets \Delta\TL_{j}+ (\delta\pi,\delta d)$
			}
		}
	    \lIf(\tcp*[f]{current best}){$\Delta\TL_j<\Delta\TL^*$} {
   			$\Delta\TL^*,y^* \gets \Delta\TL_j,y_j$
		}
	}
	$\TL, M \gets \TL + \Delta\TL^*, M \cup \{y^*\}$\tcp*[r]{use best new medoid}
	\lIf(\tcp*[f]{increase $k$}){$i=k-1$ \KwAnd $\TL_\pi>0$} {$k\gets k+1$}
}
\Return $\TL,\{m_1,...,m_k\}$\;
\label{sec:lenssen-usecase-alg1}
\end{algorithm2e}

\begin{algorithm2e}[tp]
\SetAlgoLined\DontPrintSemicolon\SetArgSty{textrm} %
\SetKwBlock{Repeat}{repeat}{}
\SetKw{ContinueOuterLoopIf}{continue outer loop if}
\SetKw{BreakOuterLoopIf}{break outer loop if}
\caption{DynSWAP: Dynamic SWAP for asymmetric sparse data}
$y_{\text{last}} \gets $ invalid\;
\lForEach {$x_o$} {compute $\nearest(o),d_\nearest(o),d_\second(o)$}
$\Delta\TL^{-m_1},\ldots,\Delta\TL^{-m_k} \gets$ compute loss change removing $m_i$\;
\While{still changing}{
    \ForEach(\label{line:sparsepam-loop1}){$y_c\notin\{m_1,\ldots,m_k\}$} {
        \BreakOuterLoopIf $y_c=y_\text{last}$ \tcp*[r]{no improvements found}
        $\Delta\TL\gets(\Delta\TL^{-m_{1}},\ldots,\Delta\TL^{-m_k})$\tcp*[r]{removal loss}
        $\Delta\TL^{+} \gets 0$\tcp*[r]{accumulator (FasterPAM)}
        \ForEach(\tcp*[f]{check neighbors (sparse)}\label{line:sparsepam-loop2}){$x_o \in N(y_c)$} {
            $d_{oc}\gets d(x_o,y_c)$\tcp*[r]{distance to candidate}
			\If(\tcp*[f]{$x_o$ not covered yet}){$d_\nearest(o) = \infty$}{ %
				$\Delta\TL^{+}\gets \Delta\TL^{+} +(-\pi(o),d_{oc})$\;\label{line:sparsepam-case1}
			}
            \ElseIf(\tcp*[f]{new nearest}){$d_{oc}<d_\nearest(o)$}{
                $\Delta\TL^{+}\gets\Delta\TL^{+} +(0,d_{oc}-d_\nearest(o))$\;\label{line:sparsepam-case2}
                \If(\tcp*[f]{no second nearest}){$d_\second(o) = \infty$}{
                	$\Delta\TL_{\nearest(o)}\gets\Delta\TL_{\nearest(o)}+(-\pi(o),d_\nearest(o))$\;
                }\Else{
                	$\Delta\TL_{\nearest(o)}\gets\Delta\TL_{\nearest(o)}+(0,d_\nearest(o)-d_\second(o))$\;
                }
            }
			\ElseIf(\tcp*[f]{no second nearest}){$d_\second(o) = \infty$}{
                $\Delta\TL_{\nearest(o)}\gets\Delta\TL_{\nearest(o)}+(-\pi(o),d_{oc})$\;\label{line:sparsepam-case3}
			}
            \ElseIf(\tcp*[f]{new second nearest}){$d_{oc}<d_\second(o)$}{
                $\Delta\TL_{\nearest(o)}\gets\Delta\TL_{\nearest(o)}+(0,d_{oc}-d_\second(o))$\;\label{line:sparsepam-case4}
			}
		}
		$i\gets \argmin(\{ \Delta\TL_i \} )$\tcp*[r]{best current medoid}
		$\Delta\TL_i\gets\Delta\TL_i+\Delta\TL^+$\tcp*[r]{add accumulator}
		\If(\tcp*[f]{eager swapping (FasterPAM)}) {$\Delta\TD_i<(0,0)$}{
			swap roles of medoid $m^*$ and non-medoid $y_c$\;
			$\TL \gets \TL+\Delta\TL_i$\;
		   	update $\nearest(o),d_\nearest(o),d_\second(o)$, $\Delta\TL^{-m_1},\ldots,\Delta\TL^{-m_k}$\;
			$y_{\text{last}}\gets y_c$\tcp*[r]{new stopping position}
    		\tcp{After each swap, try to reduce $k$:}
			\If(\tcp*[f]{$Dyn^\downarrow$}\label{line:lenssen-dyndown}) {$\min(\Delta\TL^{-m_{1}}_\pi,\ldots,\Delta\TL^{-m_k}_\pi) = 0$} {
		    	$r\gets \argmin(\{\Delta\TL^{-m_{i}}\}) $\;
		        remove medoid $m_r$\;
			    update $\nearest(o),d_\nearest(o),d_\second(o)$, $\Delta\TL^{-m_1},\ldots,\Delta\TL^{-m_k}$\;
			}
		}
	    \ElseIf(\tcp*[f]{$Dyn^\uparrow$}\label{line:lenssen-dynup}) {$\Delta\TL^+_\pi < 0$} {
        	add new medoid $y_c$ as it fixes at least one constraint \;
    		update $\nearest(o),d_\nearest(o),d_\second(o)$, $\Delta\TL^{-m_1},\ldots,\Delta\TL^{-m_k}$\;
			$y_{\text{last}}\gets y_c$\tcp*[r]{new stopping position}
    	}
    }
}
\Return {$\TL,M$}\;
\label{sec:lenssen-usecase-alg2}
\end{algorithm2e}

We introduce DynSWAP (Dynamic SWAP for asymmetric sparse data) as a dynamic SWAP algorithm
based on FasterPAM~\cite{Schubert/Rousseeuw/2019a,Schubert/Rousseeuw/2021a},
adapted to dynamically reduce~$k$, while efficiently processing asymmetric
and sparse input data. DynSWAP differs from FasterPAM's SWAP in two ways:
To dynamically change $k$ depending on the constraints, we check after
each swap whether we can reduce $k$ without violating a constraint (line~\ref{line:lenssen-dyndown}) if the current object is not suitable for swapping but reduces the number of violated constraints
if added as a new medoid (line~\ref{line:lenssen-dynup}) then we make it an additional medoid.
We deliberately choose to only reduce~$k$ if we also perform a swap, as to alternate between
optimizing the existing medoids and learning the number of clusters~$k$.
Both checks are very efficient to implement, as we already know the removal loss change for all medoids
($\Delta\TL^{-m_1},\ldots,\Delta\TL^{-m_k}$, also needed by the FastPAM improvement over PAM)
and we also have in $\Delta\TL^+$ the loss change when adding a new medoid.
We can remove the medoid~$m_i$ without breaking any constraint if the $\pi$ component is zero: $\Delta\TL^{-m_i}_\pi{=}0$,
and making the current candidate $y_j$ a new medoid is beneficial if its $\Delta\TL^+_\pi{<}0$.
Whenever adding, removing, or swapping a medoid, we need to update for all data points $x_o$ the nearest medoid
$\nearest(o)$, the distance to the nearest medoid $d_\nearest(o)$,
and the distance to the second nearest medoid $d_\second(o)$. This can be done more efficiently
by updating the previous values, exactly as in FasterPAM.
Based on this information, we can also update $\Delta\TL^{-m_1},\ldots,\Delta\TL^{-m_k}$, which is the
loss change for removing each medoid, efficiently: for each object, removing the nearest medoid
incurs a loss change of $(0,d_\second(o)-d_\nearest(o))$ if there is a second nearest medoid,
and $(\pi(o),-d_\nearest(o))$ otherwise. Removing another medoid except the nearest medoid does not
incur a loss change.

When computing the loss change for adding a new candidate medoid $y_c$,
we initialize an array $\Delta\TL$ with the removal loss of each existing medoid,
an optimization from FastPAM~\cite{Schubert/Rousseeuw/2019a}.
To avoid an inner loop over all medoids $k$, we also incorporate an idea from
FasterPAM~\cite{Schubert/Rousseeuw/2021a}, namely to accumulate the loss change
that applies to all medoids into the variable $\Delta\TL^+$.
An interesting property of $\Delta\TL^+$ is that it is the loss change
for adding a new medoid, which we use for our logic of dynamically increasing
the number of clusters, too.
We benefit from sparsity in this approach because we do not have to consider
objects that are not neighbors of the candidate $y_c$: the loss change by removing
existing medoids has already been accounted for, and as they are not reachable
from $y_c$, there is no loss change when adding the replacement medoid.
Because of this, our loop only needs to iterate over the neighbors.
For each neighbor $x_o$, we distinguish four cases: (1)~the point is currently not
yet covered, hence we gain $\pi(o)$ but incur $d(x_o,y_c)$ in line~\ref{line:sparsepam-case1},
(2)~the new medoid is closer than all existing medoids and hence we
gain $d_\nearest(o)-d(x_o,y_c)$ in line~\ref{line:sparsepam-case2}.
For the case of removing the nearest medoid,
we have already included $d_\nearest(o)$, and hence we have to cancel this out
(either with $-\pi(o)$ or $d_\second(o)$).
If the new medoid is only second nearest, and there is (3)~no previous second
nearest, only the loss of removing the nearest medoid needs to be updated in line~\ref{line:sparsepam-case3}.
If~(4)~a previous second nearest exists, but which is farther than the new medoid, we also need to adjust
the loss of removing the nearest medoid by the difference between assigning
to the new medoid instead of the previous second closest in line~\ref{line:sparsepam-case4}.
Similar case distinctions -- except for handling the case of an undefined second closest --
can already be found in FasterPAM~\cite{Schubert/Rousseeuw/2021a}.

We observe that the two loops in lines~\ref{line:sparsepam-loop1} and \ref{line:sparsepam-loop2}
iterate over all edges, hence the complexity of the procedure is $O((e+N\cdot k)\cdot i)$,
where $e$ is the number of edges and $i$ the number of iterations.
In the street network example, we can argue that $e\in O(N)$ as we scale the approach to larger networks
(as we would keep the maximum distance constant, but increase the area).
Hence, this sparse $k$-medoids version scales linearly for this application.
If we have a densely connected graph, then $e\in O(N^2)$ and the runtime matches that of standard FasterPAM.

\subsection{Experiments}
In our experiments, we expect to see a speedup compared to FasterPAM.
We also want to check how the dynamic change of $k$ works under consideration of constraints.
We have to evaluate how well the Sparse $k$-Medoids is able to find the smallest possible $k$ still meeting the constraints.
Hence, we analyze three initialization methods DynBUILD, Random, and Sparse++.
Finally, we perform a qualitative evaluation by comparing our simulations with original substation locations from OSM.
\paragraph{Data Sets}
To verify the algorithm, we need sufficiently large test data sets,
and choose constraints to obtain sparse distance matrices.
In this work, we focus on the processing and evaluation of energy grids generated by OSMOGrid.
For quality evaluation, we choose areas where many substations are documented in OSM.
\begin{figure}[t!]\centering
\includegraphics[width=1.0\textwidth]{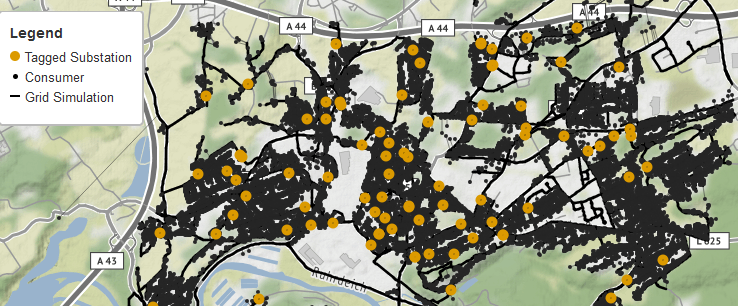}
\caption{Grid simulation and known substation locations in OSM for Witten.
The road network contains 37287 edges and 36844 nodes, $N{=}35713$ consumer and $m{=}1130$
possible places for substations. The location of 127 substations is documented in OSM,
but very likely several are missing in particular in the east.}
\label{sec:lenssen-usecase-fig3}
\end{figure}
Figure~\ref{sec:lenssen-usecase-fig3} shows a cutout of the electrical grid generated
by OSMOGrid for the city of Witten, using the 127 substation locations from OSM
(although this likely is not complete, as seen in Figure~\ref{sec:lenssen-usecase-fig3}).
We can then compare the quality of our calculated models to the model based on the real substations,
but we need to remember that there may be additional substations missing in OSM,
and that real power networks have grown historically, have to satisfy additional
constraints, and are hence not optimal either. For the purpose of generating ``realistic''
networks, it is desirable to achieve a comparable quality, without overfitting to the
example data we have.
On the dataset, we evaluate the dynamic methods for choosing~$k$.
The ``optimal'' $k$ depends on the constraints set, and thus on the sparsity of the distance matrix.
Figure~\ref{sec:lenssen-usecase-fig4} shows the smallest~$k$ to meet the constraint of
a maximum cable length in the grid.
With increasing cable length, the number of missing distances in the matrix decreases,
but the best number of substations decreases much faster.
\begin{figure}[t!]
	\center
	\begin{tikzpicture}[scale=1.0]
		  \pgfplotsset{
			scale only axis,
		}
		\begin{axis}[
		    height=35mm,
		    width=\textwidth - 2.5cm,
			axis y line*=left,
			xlabel = Maximum power cable length (m),
			xmin = 500, xmax = 6000,
			ylabel = Sparse distances (\%),
			ymin = 0, ymax = 100
			]
	\addplot[mark=square*]coordinates {
	(500, 98.53512244)
	(750, 97.14264404)
	(1000, 95.35769171)
	(1250, 93.27220034)
	(1500, 90.8997364)
	(1750, 88.24539446)
	(2000, 85.37824586)
	(2250, 82.34796746)
	(2500, 79.11184033)
	(2750, 75.80774452)
	(3000, 72.43725348)
	(3250, 68.94743441)
	(3500, 65.32985007)
	(3750, 61.72836462)
	(4000, 58.16062564)
	(4250, 54.55672176)
	(4500, 50.86955811)
	(4750, 47.02344982)
	(5000, 43.08672239)
	(5250, 39.23023667)
	(5500, 35.51582855)
	(5750, 31.99007008)
	(6000, 28.71432883)
	}; \label{plot_1_s}
		\end{axis}
	\begin{axis}[
	    height=35mm,
    	width= \linewidth - 2.5cm,
    	legend style={fill=none, draw=none, font=\small},
		axis y line*=right,
		axis x line=none,
		ylabel=Number of medoids ($k$),
		ymin = 0, ymax = 160,
		xmin = 500, xmax = 6000
		]
		\addplot[mark=x]
		coordinates{
		(500, 150)
		(600, 109)
		(700, 88)
		(800, 74)
		(1000, 56)
		(1250, 36)
		(1500, 29)
		(2000, 18)
		(2500, 15)
		(3000, 11)
		(3500, 10)
		(4000, 9)
		(4500, 8)
		(5000, 7)
		(5500, 7)
		(6000, 6)
		}; \label{plot_1_k}
	\addlegendimage{/pgfplots/refstyle=plot_1_s}\addlegendentry{best $k$ satisfying all constraints}
	\addlegendimage{/pgfplots/refstyle=plot_1_k}\addlegendentry{distances over threshold}
	\end{axis}
	\end{tikzpicture}
	\caption{Sparsity of the distance matrix depending on a maximum cable length constraint
	between consumer and substation for the simulation of Witten, and the minimum number of
	substations~$k$ for which no constraint is broken.
	}
	\label{sec:lenssen-usecase-fig4}
\end{figure}
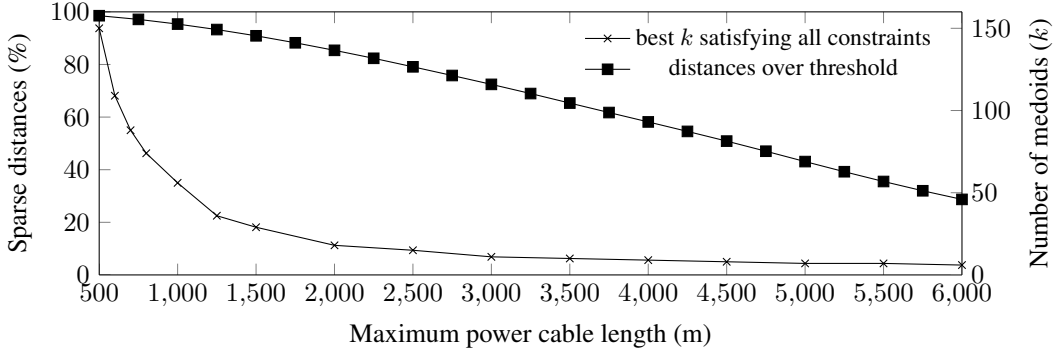

We evaluate the algorithms in the ELKI open-source toolkit~\cite{Schubert/Zimek/2019a} in Java.
For comparability, we perform all computations in ELKI and use the original implementations of FasterPAM as reference.
This way we avoid side effects caused by different implementations~\cite{Kriegel/etal/2017a}.
We run 100~restarts on an AMD EPYC~7302 processor using a single thread, and evaluate the average, maximum and minimum values.

\paragraph{Dynamic \boldmath$k$}
To evaluate the quality with a variable~$k$, we compare the solutions found by the algorithms to the best known solution of all runs.
We also compare the different initialization algorithms, and the variants of the dynamic SWAP.
We measure the solution's $k$, the runtime of initialization and SWAP, and whether the result satisfies all constraints.
Summarized results are shown in Table~\ref{sec:lenssen-usecase-tab1}. 
\begin{table}[t]\centering
	\caption{Comparison of $k$ depending on initialization and swap algorithms for generating a grid for Witten.
	All results are averaged over 100 restarts.}
	\label{sec:lenssen-usecase-tab1}
    \setlength{\tabcolsep}{6pt}
	\begin{tabular}{l@{\;}lr@{}lr@{}lr@{}lr@{}lr@{}lr@{}lr}
		\toprule
		\multicolumn{2}{c}{\bf Algorithm} & \multicolumn{4}{c}{\bf\boldmath$k$ after} & \multicolumn{4}{c}{\bf runtime in ms} & \multicolumn{2}{c}{\bf medoid} & \multirow{2}{*}{\bf success}\\
		Init. & SWAP & \multicolumn{2}{c}{Init} & \multicolumn{2}{c}{SWAP} & \multicolumn{2}{c}{Init} & \multicolumn{2}{c}{SWAP} & \multicolumn{2}{c}{\bf changes}  &\\
		\midrule
		\multirow{2}{*}{Random\textsuperscript{5}} & Dyn$^{\downarrow}$ & 56& & 56& &\bf 0&\bf.2 & 6919&.2 & 205&.3 & 0\% \\
		& Dyn$^{\downarrow\uparrow}$ & 56& & 77&.9 &\bf 0&\bf.2 & 7862&.3 & 287&.9 & 100\% \\
	    \multirow{2}{*}{Random\textsuperscript{10}} & Dyn$^{\downarrow}$ & 113& & 82&.8 & \bf 0&\bf .1 & 11578&.8 & 354&.9 & 100\% \\
		& Dyn$^{\downarrow\uparrow}$ & 113& & 82&.5 & \bf 0&\bf .1 & 11044&.4 & 372&.9 & 100\% \\
		\multirow{2}{*}{Sparse++} & Dyn$^{\downarrow}$ & 182&.3 & 80&.3 & 16&.3 & 8632&.2 & 269&.2 & 100\% \\
		& Dyn$^{\downarrow\uparrow}$ & 182&.2 & 80&.6 & 3&.6 & 8640&.3 & 268&.4 & 100\% \\
		\multirow{2}{*}{DynBUILD} & Dyn$^{\downarrow}$ & 93& &\bf 76&\bf.3 & 389&.9 & \bf 4296&\bf .3 & 141&.9 & 100\% \\
		& Dyn$^{\downarrow\uparrow}$ & 93& & \bf 76&\bf .2 & 394&.6 & 4320&.3 & 141&.4 & 100\% \\
		\bottomrule
	\end{tabular}
\end{table}
Only Random$^{5}$ initialization without dynamic increase of~$k$ fails to satisfy constraints.
This was to be expected, because it only uses 5\% of the possible substations as medoids,
but there does not appear to be a solution with just this many clusters.
Because DynBUILD is deterministic, it always produced
$k{=}93$ clusters after initialization. After the SWAP phase, the average $k$ was $76.2$,
which is $2.2$ more than the best known $k{=}74$ (we iterate in a randomized order in SWAP to avoid dependence on the input data order).
Since the initial solution already satisfied all constraints, and SWAP preserves this property, $k$ can only decrease.
Among the random initializations Random\textsuperscript{5} with DynSWAP$^{\downarrow\uparrow}$ found the best results on average.
With $k {=} 77.9$ after the SWAP phase, the number of stations is on average $3.9\times$ larger than the best~$k$.
Finally, the SWAP after DynBUILD needs on average 48.4\% of the runtime of the SWAP after a random initialization.
Due to random initialization, the average number of medoid changes during the SWAP increases significantly from 141 to 310,
showing that DynBUILD finds superior starting conditions than both random sampling and Sparse++. 

\paragraph{Runtime Speedup}
In order to evaluate the runtime of the different methods, we perform experiments
for varying constraints and values of~$k$.
\begin{figure}[b!]\centering
	\begin{tikzpicture}[scale=1.0,font=\small]
		\pgfplotsset{
			scale only axis,
		}
		\begin{axis}[
		    legend style={at={(0.63,0.7)},anchor=north,draw=none,inner sep=0,font=\scriptsize},
		    legend cell align={left},legend columns=2,
			height=4.0cm,
    	    width= \textwidth - 2.0cm,
			xlabel = Best known number of clusters k,
			xmin = 36, xmax = 150,
			ylabel = {runtime (ms, log scale)},
			ymin = 3000, ymax = 300000,
			ymode=log
			]
			\addplot[Pa2, mark=triangle*]coordinates {
			(36, 6268.2)
			(56, 5588.0)
			(74, 5546.2)
			(88, 4853.9)
			(150, 3796.0)
			}; \label{plot_1b_s}
			\addplot[Pa3, mark=diamond*]coordinates {
			(36, 13164.3)
			(56, 9414.5)
			(74, 7724.1)
			(88, 7380.4)
			(150, 7847.0)
			}; \label{plot_5r_s}
			\addplot[Pa4, mark=star]coordinates {
			(36, 25820.4)
			(56, 17628.7)
			(74, 11597.6)
			(88, 8482.2)
			(150, 6393.5)
			}; \label{plot_1r_s}
			\addplot[Pa1, mark=star]coordinates {
			(36, 13164.7)
			(56, 10190.9)
			(74, 8741.9)
			(88, 9236.7)
			(150, 8279.4)
			}; \label{plot_1p_s}
			\addplot[Pa5, mark=square*]coordinates {
			(36, 61238.4)
			(56, 91827.7)
			(74, 108307.8)
			(88, 120855.3)
			(150, 272387.8)
			}; \label{plot_0b_s}
			\addlegendimage{/pgfplots/refstyle=plot_1b_s}\addlegendentry{DynBUILD - Dyn$^{\downarrow}$SWAP}
			\addlegendimage{/pgfplots/refstyle=plot_5r_s}\addlegendentry{Random\textsuperscript{5} - Dyn$^{\downarrow \uparrow}$SWAP}
			\addlegendimage{/pgfplots/refstyle=plot_1r_s}\addlegendentry{Random\textsuperscript{10} - Dyn$^{\downarrow \uparrow}$SWAP}
			\addlegendimage{/pgfplots/refstyle=plot_1p_s}\addlegendentry{Sparse++ - Dyn$^{\downarrow \uparrow}$SWAP}
			\addlegendimage{/pgfplots/refstyle=plot_0b_s}\addlegendentry{Random - FasterPAM}
		\end{axis}
	\end{tikzpicture}
	\begin{tikzpicture}[scale=1.0, font=\small]
		\pgfplotsset{
			scale only axis,
		}
		\begin{axis}[
		    legend style={at={(0.4,0.97)},anchor=north,draw=none,inner sep=0,font=\scriptsize},
		    legend cell align={left},legend columns=2,
			height=4.0cm,
    	    width= \textwidth - 2.0cm,
			xlabel = Best known number of clusters k,
			xmin = 36, xmax = 150,
			ylabel = Excess clusters k,
			ymin = 0, ymax = 15
			]
			\addplot[Pa2, mark=square*]coordinates {
			(36, 2.2)
			(56, 1.0)
			(74, 2.1)
			(88, 2.6)
			(150, 2.9)
			}; \label{plot_1b_k}
			\addplot[Pa3, mark=triangle*]coordinates {
			(36, 8.5)
			(56, 3.9)
			(74, 4.11)
			(88, 4.4)
			(150, 4.0)
			}; \label{plot_5r_k}
		    \addplot[Pa4, mark=diamond*]coordinates {
			(36, 9.8)
			(56, 6.4)
			(74, 8.6)
			(88, 10.9)
			(150, 5.7)
			}; \label{plot_1r_k}
			\addplot[Pa1, mark=star]coordinates {
			(36, 6.8)
			(56, 4.5)
			(74, 6.5)
			(88, 9.0)
			(150, 7.4)
			}; \label{plot_1p_k}
			\addlegendimage{/pgfplots/refstyle=plot_1b_k}\addlegendentry{DynBUILD - Dyn$^{\downarrow}$SWAP}
			\addlegendimage{/pgfplots/refstyle=plot_5r_k}\addlegendentry{Random\textsuperscript{5} - Dyn$^{\downarrow \uparrow}$SWAP}
			\addlegendimage{/pgfplots/refstyle=plot_1r_k}\addlegendentry{Random\textsuperscript{10} - Dyn$^{\downarrow \uparrow}$SWAP}
			\addlegendimage{/pgfplots/refstyle=plot_1p_k}\addlegendentry{Sparse++ - Dyn$^{\downarrow \uparrow}$SWAP}
		\end{axis}
	\end{tikzpicture}
	\caption{Runtime of the initialization and SWAP for DynBUILD, Random\textsuperscript{5}, Random\textsuperscript{10} and Sparse++ initialization depending
	on the best number of $k$ for the simulation of the grid of Witten. For reference, the random initialization and SWAP runtime of the FasterPAM implementation is included, where the $k$ chosen is the best one we know.
	The best $k$ is controlled indirectly by the constraints set, see Figure~\ref{sec:lenssen-usecase-fig4}.
	In addition to the runtime, the deviation from the best known $k$ after SWAP is also shown.}
	\label{sec:lenssen-usecase-fig5}
\end{figure}
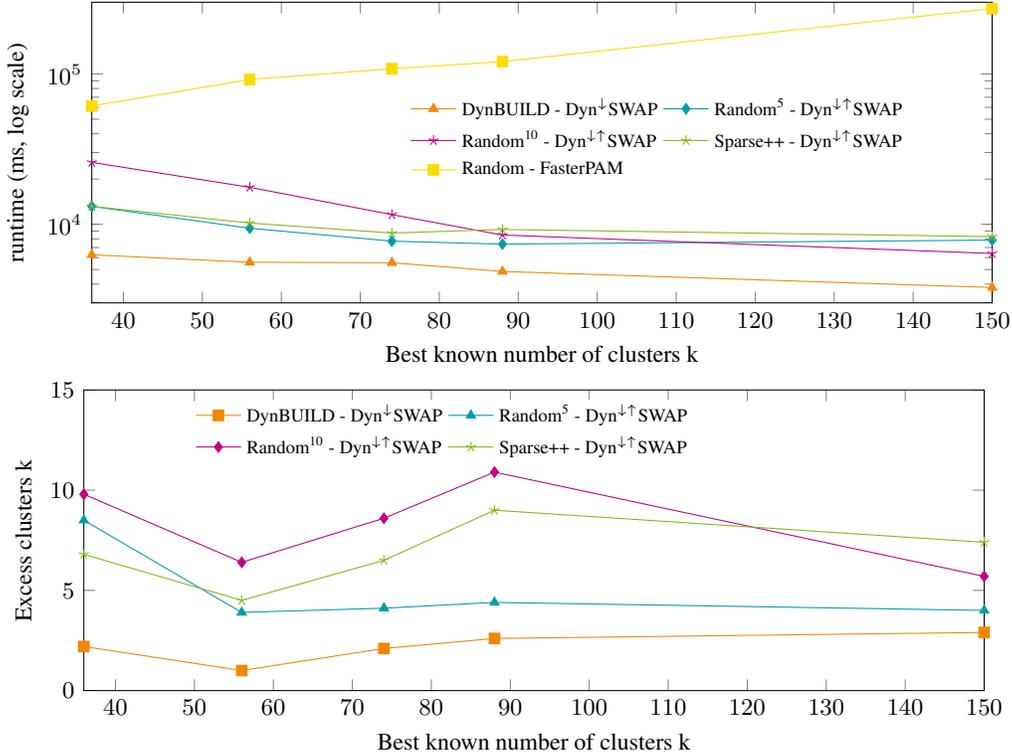
Figure~\ref{sec:lenssen-usecase-fig5} shows the total runtime (initialization and SWAP) for
DynBUILD, Random\textsuperscript{5}, Random\textsuperscript{10}, and Sparse++ initialization.
We again use the Witten data set and choose the distance constraint such that all methods can achieve the desired~$k$.
We compare the runtime with the FasterPAM implementation with a random initialization (as recommended for FasterPAM).
For DynBUILD we evaluate the SWAP with dynamic decrease of~$k$ (Dyn$^{\downarrow}$)
and for all random initialization the SWAP with both dynamic increase and decrease of~$k$ (Dyn$^{\downarrow \uparrow}$). We use a log scale on this plot because of the huge differences: the sparsity optimized DynSWAP over all initializations on average uses only 7\% of the runtime of the original FasterPAM with random initialization, the DynBUILD and DynSWAP$^{\downarrow}$
combination on average even only uses~4\%. This was expected as FasterPAM has to process the much larger dense matrix.
With increasing $k$, we can use a more sparse matrix here, which is the reason why the DynSWAP
approaches become faster while FasterPAM becomes slower due to the higher number of clusters.
The various random initializations differ only slightly in runtime,
but require on average about twice as long as DynBUILD.
In addition to the fast runtime, DynBUILD with Dyn$^{\downarrow}$SWAP also produces the
lowest number of excess clusters compared to the best known~$k$ with an average of 2.2 stations more than the best known $k=74$.

\paragraph{Quality}
\begin{table}[b!]\centering
	\caption{Comparison of the loss of the grid with 127 tagged substations with the calculated loss for $k$ = 127.
	All results are given as average values of 100 restarts. All constraints were satisfied after the SWAP phase.}
	\label{sec:lenssen-usecase-tab2}
	\setlength{\tabcolsep}{4pt}
	\begin{tabular}{llr@{}lr@{}lr@{}lr@{}lr@{}lr@{}lr@{}lr@{}lr@{}l}
		\toprule
		\multicolumn{2}{c}{\multirow{2}{*}{\textbf{Algorithm}}} & \multicolumn{8}{c}{{\bf loss after SWAP} $\times 10^7$} &  \multicolumn{4}{c}{\multirow{2}{*}{\textbf{runtime in ms}}} & \multicolumn{2}{c}{\multirow{2}{*}{\textbf{medoid}}}  \\
		& & \multicolumn{4}{c}{$\varnothing$} & \multicolumn{4}{c}{min}  & && && \multicolumn{2}{c}{\multirow{2}{*}{\textbf{changes}}}  \\
		Init. & SWAP & \multicolumn{2}{c}{d} & \multicolumn{2}{c}{$\pi$} & \multicolumn{2}{c}{d} & \multicolumn{2}{c}{$\pi$} & \multicolumn{2}{c}{Init} & \multicolumn{2}{c}{SWAP} &&  \\
		\midrule
		Random\textsuperscript{5} & Dyn$^{\downarrow\uparrow}$ & 1&.3155 & \textbf{0}& & \textbf{1}&\textbf{.3083} & \textbf{0}& & \textbf{0}&\textbf{.1} & 12418&.6 & 469&.6   \\
		Random\textsuperscript{10} & Dyn$^{\downarrow\uparrow}$ & \textbf{1}&\textbf{.3149} & \textbf{0}& & 1&.3085 & \textbf{0}& & \textbf{0}&\textbf{.1} & 10395&.5 & 339&.8   \\
		Sparse++ & Dyn$^{\downarrow\uparrow}$ & \textbf{1}&\textbf{.3149} & \textbf{0}& & \textbf{1}&\textbf{.3083} & \textbf{0}& & 16&.4 & 10796&.5 & 279&.4   \\
		DynBUILD & Dyn$^{\downarrow}$ & 1&.3171 & \textbf{0}& & 1&.3092 & \textbf{0}& & 525&.6 & \textbf{9225}&\textbf{.4} & 294&.9   \\
		\midrule
		\multicolumn{2}{l}{\textbf{tagged substations}} & 1&.86 & \textbf{0} & & && & & \\
		\bottomrule
	\end{tabular}
\end{table}

\begin{figure}[t!]\centering
\includegraphics[width=1.0\textwidth]{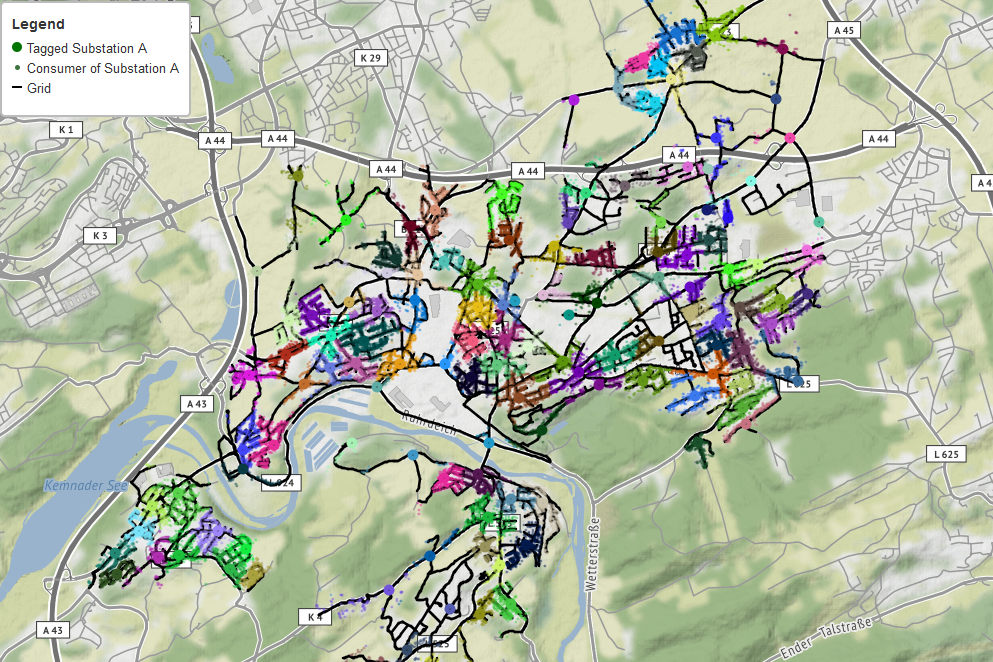}
\caption{Simulation of an electrical grid based on OSM data of Witten.
127 substations calculated with Sparse++ and Dyn$^{\downarrow\uparrow}$SWAP from Table~\ref{sec:lenssen-usecase-tab2}.
All consumers are color-coded to their nearest sub-station.}
\label{sec:lenssen-usecase-fig6}
\end{figure}
In order to evaluate the resulting quality, we compare the optimized substation locations to the
substations tagged in the OSM (however, these are likely incomplete).
Table~\ref{sec:lenssen-usecase-tab2} shows the results for a target $k {=} 127$
compared to the loss of the 127 tagged substations, shown in Figure~\ref{sec:lenssen-usecase-fig3}.
Sparse++ and Random\textsuperscript{10} initialization with Dyn$^{\downarrow\uparrow}$SWAP results in the lowest loss with 1.3149$\times 10^7$
and is 29\% lower than the loss of the tagged substations; but the quality difference between the
randomized initializations is not significant (the SWAP does a good enough job to always reach a good solution);
the main difference here is in the runtimes, where the strategy to sample more centers than necessary, then decrease,
seems to be superior to the others. 
The minimum loss over 100 restarts is obtained with Sparse++ and Random\textsuperscript{5} with 1.3083$\times 10^7$.
DynBUILD initialization with Dyn$^{\downarrow}$SWAP finds a slightly higher
loss of 1.3271$\times 10^7$, but yields the fastest total runtime with 9551.0~ms,
despite using the slowest initialization by far.

All methods found significantly better solutions (1.31 vs.{} 1.86) than
the ``gold standard'' solution given by the OSM tags. This had to be expected
because of supposedly missing tags, but also because the real grids were grown over time
and the substations were built one by one, and not automatically optimized.
In our simulation, we have complete information about the grid structure and can thus
calculate an optimal substation distribution (green field planning),
that cannot be realistically achieved in practice,
because the existing power network cannot simply be replaced and has obey
additional constraints. Nevertheless, the resulting networks can be
useful for simulating power networks in different scenarios, for example
when investigating the effect of significantly expanding the charging infrastructure
for electric cars.

\subsection{Outlook}
In the experiments, we focused on the specific use case of energy grid simulation.
Besides optimizing the FasterPAM approach for sparse problems, we have begun
working on automatically finding the parameter $k$ as part of the optimization problem.
For this, we combined two losses in our loss function: one corresponding to the cost
of poorly handled locations (which could also be outliers), and the other part being
the classic $k$-medoids problem. It would be easy to incorporate an additional cost term
to the opening or closing of locations, and to weigh these costs differently.
In this experiment, we used a strict requirement to cover all locations (i.e., $\pi\rightarrow\infty$),
but using a smaller weight may yield interesting approximations.

So far, we have considered the problem of optimal substation positions without a maximum capacity of substations.
In reality, there is a maximum load that can be served by a substation.
In densely populated areas, we hence may need more substations.
This results in a capacitated facility location problem,
which contains such an additional capacity constraint,
and is worth exploring in future work.

\let\showmark\undefined

\fi

\egroup

\printbibliography
\end{document}